\documentclass[runningheads]{llncs}
\usepackage{graphicx}
\usepackage{amsmath,amssymb} 
\usepackage{color}
\usepackage{epsfig}
\usepackage{multirow, booktabs}
\usepackage{enumitem}
\usepackage{caption}
\usepackage{amsmath,bm}
\usepackage{makecell}
\usepackage{authblk}
\usepackage[symbol*]{footmisc}
\usepackage[dvipsnames]{xcolor}

\def\@fnsymbol#1{\ensuremath{\ifcase#1\or *\or \dagger\or \ddagger\or
   \mathsection\or \mathparagraph\or \|\or **\or \dagger\dagger
   \or \ddagger\ddagger \else\@ctrerr\fi}}

\begin{document}

\title{Matchable Image Retrieval by Learning from Surface Reconstruction} 
\titlerunning{Matchable Image Retrieval by Learning from Surface Reconstruction} 


\author{Tianwei Shen\thanks{denotes equal contributions.}\inst{1}\orcidID{0000-0002-3290-2258} \and Zixin Luo$^{*}$\inst{1}\orcidID{0000-0001-6946-2826} \and ~~~~~~~~~~~~~~Lei Zhou\inst{1}\orcidID{0000-0003-4988-5084} \and Runze Zhang\inst{1}\orcidID{0000-0001-9698-0178} \and ~~~~~~~~~~~~~~~Siyu Zhu\inst{1}\orcidID{0000-0003-0293-0044} \and Tian Fang\thanks{Tian Fang is the corresponding author.}\inst{2}\orcidID{0000-0002-5871-3455} \and ~~~~~~~~~~~~~~~~Long Quan\inst{1}\orcidID{0000-0001-8148-1771}}

%

\authorrunning{T. Shen, Z. Luo et al.} 

\institute{Hong Kong University of Science and Technology, Hong Kong, China
\\ \email{\{tshenaa,zluoag,lzhouai,rzhangaj,szhu,quan\}@cse.ust.hk}\\
 \and
Shenzhen Zhuke Innovation Technology (Altizure), Shenzhen, China\\ \email{\{fangtian\}@altizure.com}}

\newcommand{\sfmDatasetName}{Geometric Learning with 3D Reconstruction (GL3D)}
\newcommand{\testSetNum}{40 }
\newcommand{\trainSetNum}{338 }
\newcommand{\totalSetNum}{378 }
\newcommand{\norm}[1]{\left\lVert #1 \right\rVert}

\maketitle

\begin{abstract}
Convolutional Neural Networks (CNNs) have achieved superior performance on object image retrieval, while Bag-of-Words (BoW) models with handcrafted local features still dominate the retrieval of overlapping images in 3D reconstruction. In this paper, we narrow down this gap by presenting an efficient CNN-based method to retrieve images with overlaps, which we refer to as the \textit{matchable image retrieval} problem. Different from previous methods that generates training data based on sparse reconstruction, we create a large-scale image database with rich 3D geometrics and exploit information from surface reconstruction to obtain fine-grained training data. We propose a batched triplet-based loss function combined with mesh re-projection to effectively learn the CNN representation. The proposed method significantly accelerates the image retrieval process in 3D reconstruction and outperforms the state-of-the-art CNN-based and BoW methods for matchable image retrieval. The code and data are available at \url{{\color{VioletRed} https://github.com/hlzz/mirror}}.
\end{abstract}

\section{Introduction}
Generic image retrieval is widely employed in practical Structure-from-Motion (SfM)~\cite{agarwal2011building,moulon2013global,sweeney2015optimizing,schonberger2016structure} and visual simultaneous localization and mapping (SLAM)~\cite{mur2015orb} systems to accelerate the image matching process or identify possible closed loops. 
Until recently, the preferred image retrieval techniques used in SfM are largely variants of the Bag-of-Words (BoW) models~\cite{sivic2003video,nister2006scalable}, despite the fact that CNN-based approaches~\cite{kalantidis2015cross,tolias2015particular,radenovic2016cnn,iscen2017efficient} have shown superior efficiency and scalability for particular object retrieval.

This discrepancy can be explained by the difference between \emph{semantic similarity} and \emph{geometric similarity}. For SfM tasks, geometric overlaps among images (geometric similarity), rather than information about object categories (semantic similarity), are required for later reliable image matching. We refer to this specific type of image retrieval task as \emph{matchable image retrieval}, the goal of which is to find images with large overlaps. 
Two images are overlapped if they include the same area of the viewed objects or scenes.
In this scenario, BoW models based on local descriptors are more robust since they serve as predictors~\cite{havlena2014vocmatch} for how well the local descriptors can be matched.
However, neither BoW models nor CNN-based methods perfectly solve the matchable image retrieval problem.
On the one hand, BoW models generally have limited scalability as the efficiency and accuracy drop quickly with the increase of data. 
CNN-based methods, on the other hand, offer efficient and scalable solutions by compact global image representations distilled from intermediate feature maps, yet they lack the ability to identify regional discriminations and local information. 
This problem has long been overlooked because nearly all of these CNN-based methods are evaluated on object retrieval datasets such as Oxford5k~\cite{philbin2007object} and Paris6k~\cite{philbin2008lost}, in which images are organized by semantic similarity rather than geometric overlaps.

However, in a typical SfM scene (Fig.~\ref{fig:dataset_view}) consisting of overlapping images with weak semantics, current CNN-based methods are worse than BoW models because they fail to render a fine-grained ranking with respect to scene overlaps. That is probably the reason why stable SfM~\cite{agarwal2011building,moulon2013global,sweeney2015optimizing,schonberger2016structure} and SLAM~\cite{mur2015orb} solutions still adopt BoW models for matchable image retrieval.
CNN-based methods should be employed because of its superior efficiency and scalability, and its previous success in object retrieval tasks. However, several problems should be addressed to get rid of the above flaws. 
First, we are in need of a large-scale SfM database to avoid the data bias in previous evaluations. Second, information about geometric relationships between images should be further exploited to better encode local information. Several methods such as~\cite{radenovic2016cnn} have attempted to do so but stayed in the SfM level instead of using dense correspondences. Third, the training process should be made more efficient to cope with big data.

In this paper, we present an efficient CNN-based method for matchable image retrieval that utilizes rich geometric context mined from densely reconstructed structures, namely mesh re-projection and overlap masks. Moreover, local information is taken good care of with a post-processing step that exploits regional matching. In summary, our contributions are threefold:
\smallskip\begin{itemize}[leftmargin=*, noitemsep, topsep=0pt]
    \item We present \emph{\sfmDatasetName}, a large-scale database for 3D reconstruction and geometry-related learning problems, which contains \totalSetNum different datasets with full coverage of the scenes.
    
    \item We make use of the dense correspondences mined from 3D reconstruction to develop an automatic pipeline for ground-truth data generation, which results in fine-grained training data with respect to scene overlaps.
    
    \item We propose \textit{mask triplet loss (MTL)} with in-batch mining which utilizes the well-annotated training data combined with regional information to accelerate the training of matchable image retrieval.
\end{itemize}
\smallskip

\section{Related Works}

\noindent\textbf{Local descriptor based methods.} In the 3D modeling of city-scale imagery, either pairwise image matching or point-cloud matching~\cite{zhou2017progressive,zhou2018learning} often take a majority of computation. 
Since the seminal work of reconstructing Internet imagery~\cite{agarwal2011building}, object retrieval techniques have been widely adopted in a series of SfM systems~\cite{moulon2013global,sweeney2015optimizing,shen2016graph,zhu2017parallel,zhu2018very}. 
As a successful BoW model, vocabulary tree~\cite{nister2006scalable} has become indispensable in large-scale SfM, which can be regarded as a preemptive filtering step in which local descriptors vote for images that share scene overlaps. 
Later works focus on decreasing quantization errors~\cite{jegou2008hamming,philbin2008lost,li2015pairwise}, applying post-processing steps~\cite{chum2011total} and scaling up object retrieval by aggregating local features into compact representations.
To address the very large retrieval problem, VLAD~\cite{jegou2010aggregating} was designed to be a low dimensional compact code while still preserving good performance.

\noindent\textbf{CNN methods.}
Different from BoW models, CNN-based image retrieval approaches mostly rely on global information. 
Generic deep descriptors extracted from deep convolutional neural network models are proved to be good image representations on a series of vision tasks including object retrieval. 
Babenko et al.~\cite{babenko2015aggregating} firstly propose a sum-pooling aggregation method utilizing a centering prior, with the knowledge that objects of interest tend to be located close to the center of images. 
This is not satisfied when finding similarity pairs in terms of region overlaps.
Kalantidis et al.~\cite{kalantidis2015cross} later propose a feature aggregation method based on cross-dimensional weighting.
It analyses the spatial weighting and the channel weighting strategies that can boost saliency and distinctiveness of the visual patterns respectively. 
In parallel, Tolias et al.~\cite{tolias2015particular} propose R-MAC (regional maximum activations of convolutions), which utilizes regional information to boost the performance. Gordo et al.~\cite{gordo2017end} replace the rigid grid with a learned region proposal network (RPN).
All of the above methods, evaluations and assumptions are based on images with salient semantic regions like houses or landscapes. 
In 3D reconstruction, however, many images in urban datasets or aerial imageries merely serve as bridges to connect partial scenes, with fragmental, discontinuous or even no semantically meaningful regions. 

In terms of the network architectures, our method is most similar to Wang et al.~\cite{wang2014learning} and Schroff et al.~\cite{schroff2015facenet}. 
Both methods employ triplet loss to learn a similarity-based embedding. 
The first work uses a triplet-based hinge loss to characterize fine-grained image similarity while the second is proposed to solve face recognition problem at scale. 
Melekhov et al.~\cite{melekhovsiamese} also tackle the similar whole-image matching problem while they use the 2-channel network~\cite{chopra2005learning}, but do not go deeper into 3D reconstruction.


\section{The GL3D Benchmark Dataset}
\begin{figure}[t]
    \centering 
    \includegraphics[width=\textwidth]{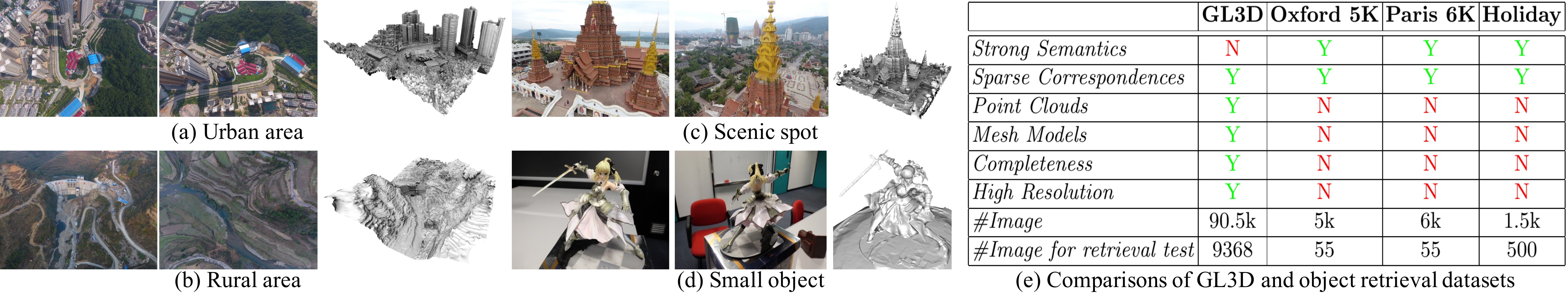}
    \caption{(a)(b)(c)(d) show different types of scenes in the GL3D dataset, with the mesh models on the right for generating training samples. (e) Compared with existing object retrieval datasets, GL3D offers high-resolution and complete views for 379 different scenes from which 2D feature matches (sparse correspondence), 2D-3D correspondences, point clouds, and mesh models can be established}
    \label{fig:dataset_view}
\end{figure}

We create a database, \emph{\sfmDatasetName}, containing 90,590 high-resolution images in \totalSetNum different scenes. Each scene contains 50 to 1,000 images with large geometric overlaps, covering urban, rural area, or scenic spots captured by drones from multiple scales and perspectives. It also contains small objects to enrich the data diversity.
Fig.~\ref{fig:dataset_view} gives an overview of various scenes in GL3D and their corresponding 3D models. We randomly select \trainSetNum datasets (81,222 images) and run 3D reconstruction pipeline (SfM $\rightarrow$ dense reconstruction $\rightarrow$ mesh reconstruction) for training sample generation as described in Section~\ref{sec:data_generation}. To generate the 3D models, we use the incremental SfM method from~\cite{zhu2018very} and the multiview stereo with surface reconstruction method from~\cite{li2016efficient}. The testing is carried out on the other \testSetNum datasets with 9,368 images as queries, which allows a thorough evaluation for the matchable image retrieval compared with only 55 queries in Oxford5k~\cite{philbin2007object}.

GL3D is tailored for geometry-related problems and offers rich 3D context information such as feature-track correspondences, camera poses, point cloud data and mesh models. Therefore, it has intrinsic difference with other existing object retrieval datasets such as Oxford5k~\cite{philbin2007object}, Paris6k~\cite{philbin2008lost} and Holiday~\cite{jegou2008hamming}. The comparisons can be characterized in the following perspectives:

\noindent\textbf{Full coverage.} Each dataset has full coverage of the scene, which is the major difference of GL3D from previous crowd-sourced datasets~\cite{radenovic2016cnn}. Existing object retrieval datasets usually have uneven samples of the same landmark, while GL3D are organized by densely connected images from different views.

\noindent\textbf{Weak semantics.} Existing object retrieval datasets mainly contain semantically meaningful landmark buildings with intact objects. The superior CNN performance trained on object classification task is therefore suitable to be transferred to object retrieval.
In contrast, GL3D has weak semantics because images only capture part of the objects or scenes without definite semantic meanings.
Some query images even have texture-less patterns like lawns and rivers, which is not common in the datasets for particular object retrieval~\cite{philbin2007object,philbin2008lost,jegou2008hamming}.

\noindent\textbf{Rich geometric context.} Since images are densely connected and have full coverage of the scenes, not only two-view feature matches, but also accurate geometric computations such as camera poses, point clouds and mesh models can be derived. Therefore, we can measure the degree of scene overlaps between images pairs from accurate mesh re-projection. This results in the proposed fine-grained ground-truth generation scheme.

GL3D is not only limited to the matchable image retrieval problem. With various geometric computations such as feature matching, camera poses, and mesh models, it is also beneficial for other geometric learning problems. For the task of matchable image retrieval, we will design and present an automatic pipeline to generate well-annotated data as described in Section~\ref{sec:data_generation}.

\section{Method}

\subsection{Problem Formulation}
Given a set of $N$ images $\{\mathcal{I}_i\}$ with geometric overlaps, we aim to find a rank set $\mathcal{S}_i$ for each image in $\{\mathcal{I}_i\}$. 
In the rank set $\mathcal{S}_i = \{\mathcal{I}_{i_1}, \mathcal{I}_{i_2}, \dots, \mathcal{I}_{i_N}\}$, a natural ordering exists $\mathcal{O}_i = (i_1, i_2, \dots, i_N)$ representing the similarity in terms of geometric overlaps between $\mathcal{I}_i$ and database images. 
To find these rank sets, one typical approach is to first map image features onto a space with lower dimension via an embedding function $f(\mathcal{I}_i)$~\cite{tolias2015particular,babenko2015aggregating,chopra2005learning,ustinova2016learning}. 
Then a similarity measurement $D(f(\mathcal{I}_i), f(\mathcal{I}_j))$ is computed and similar items are ranked by this similarity score from low to high. 
The similarity measurement is typically defined as the  $L2$ distance between two normalized feature vectors:
\begin{equation}
D(f(\mathcal{I}_i), f(\mathcal{I}_j)) = \norm{\frac{f(\mathcal{I}_i)}{\norm{f(\mathcal{I}_i)}} - \frac{f(\mathcal{I}_j)}{\norm{f(\mathcal{I}_j)}} }_2
\end{equation}

The most crucial part in this learning framework is to find the embedding function $f(\cdot)$. 
In this work, we resort to deep CNNs for embedding learning. 
Our objective is to train a neural network that can differentiate the degree of scene overlaps between pairs of images. 

\subsection{Network Architecture}\label{sec:net_arch}
\begin{figure}[]
    \centering 
    \includegraphics[width=\textwidth]{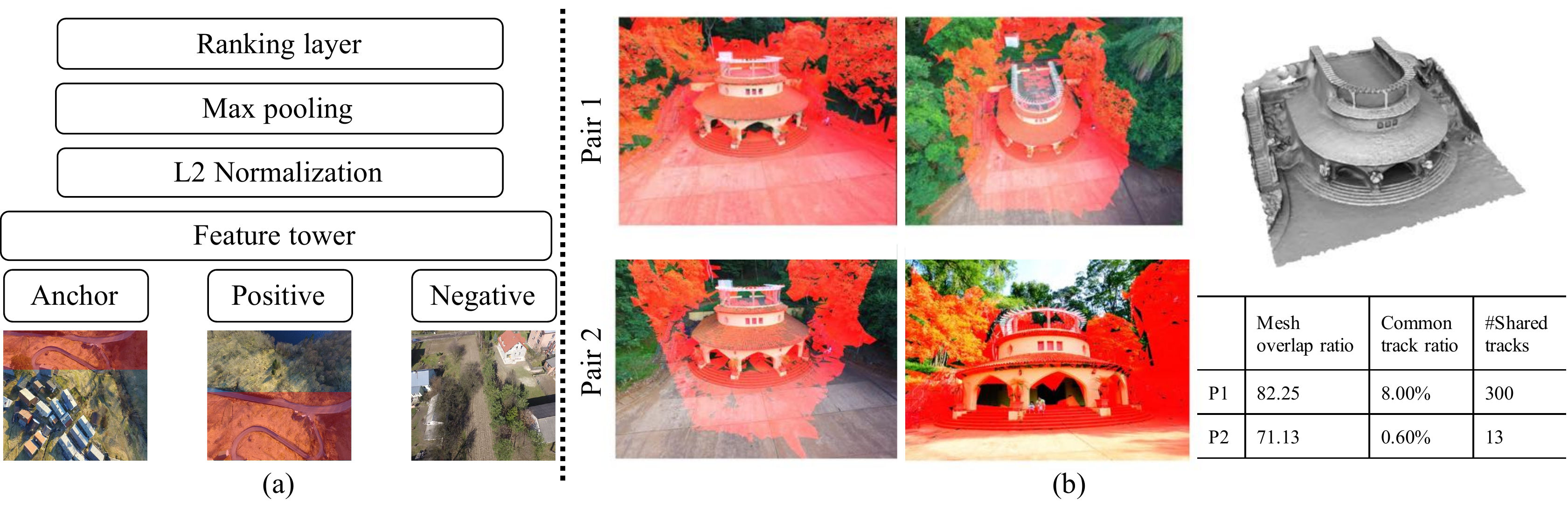}
    \caption{(a) The proposed network architecture: Three-branch feature embedding towers conjoined by the ranking layer. The anchor-positive image pair is associated with a pair of down-sampled corresponding masks (the red regions), used in the loss computation (see Equation~\ref{eq:tle}) (b) The limitation of \emph{common track ratio}. The image pair 2 survives in \emph{mesh overlap ratio} but fails in \emph{common track ratio} due to large perspective and scale changes. The {\color{red}red} masks generated by mesh re-projection indicate the overlapping regions. The 3D model and overlap statistics are presented on the right}
    \label{fig:net_arch}
\end{figure}

We adopt three-branch networks as shown in Fig.~\ref{fig:net_arch}, with \textit{(anchor, positive, negative)} image triplets (denoted by $(\mathcal{I}_a, \mathcal{I}_p, \mathcal{I}_n)$) as inputs. The core of this learning method is to minimize the distance of similar image pairs and maximize the distance of dissimilar pairs to some margin.
The embedding function $f(\cdot)$ is learned in three feature towers with shared parameters, which can be implemented with any commonly-used CNNs~\cite{simonyan2014very,szegedy2015going}. Different components in the networks are described in detail as follows:

\noindent\textbf{Feature tower.}
The three feature towers share the same parameters during training, following the essence of triplet loss. Feature tower can be fine-tuned from the widely adopted networks such as VGG~\cite{simonyan2014very} or GoogLeNet~\cite{szegedy2015going}. Though the classical networks often come with a fully-connected (FC) layer for classification, FC layers often do not work well for image retrieval tasks~\cite{tolias2015particular}. In addition, FC layers are often removed for testing because we would like the input image to be arbitrary size. Therefore, we make the feature tower to be fully convolutional. The feature vectors are composed by first applying pooling on each feature map and then $L2$ normalization across channels.

\noindent\textbf{Pooling layer.} We use max pooling to aggregate feature maps into a feature vector. Max pooling has the nice property of translation invariance and widely adopted by previous CNN image retrieval works~\cite{tolias2015particular,radenovic2016cnn,iscen2017efficient,gordo2017end}.

\noindent\textbf{Loss function.}
We use the widely adopted triplet-based loss layer for this learning-to-rank problem.
Although pairwise losses such as the contrastive loss~\cite{chopra2005learning} based on Siamese architecture~\cite{melekhovsiamese,radenovic2016cnn} are also feasible, triplet-based losses are typically favored to avoid overfitting as they care about the relative ordering rather than the absolute distance~\cite{ustinova2016learning}.
We conjoin each feature tower to the ranking layer and evaluate the hinge loss of a triplet.

\subsection{Fine-grained Training Data Generation}\label{sec:data_generation}

\begin{figure}[]
    \centering 
    \includegraphics[width=\textwidth]{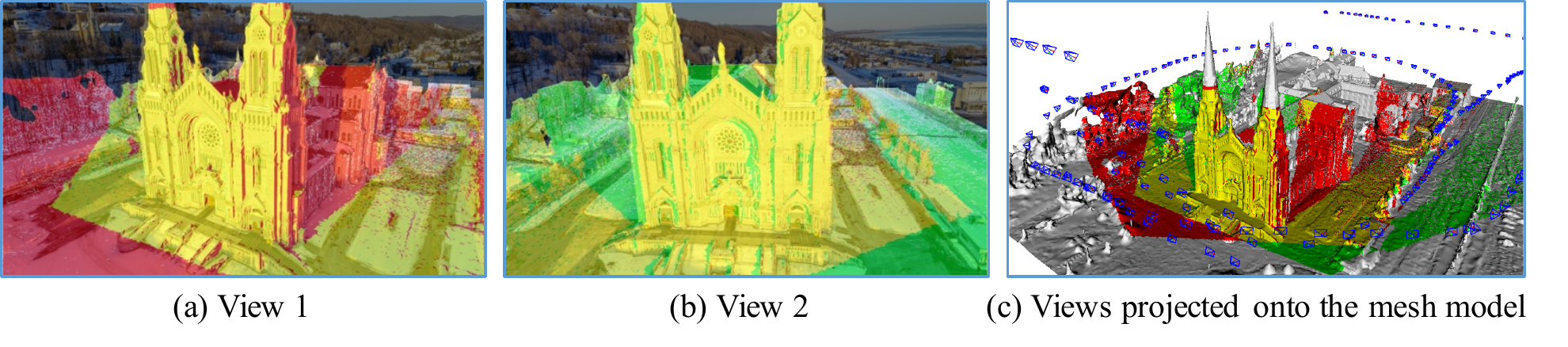}
    \caption{The automatic training data generation pipeline. The yellow region covers all the triangles that are seen by both (a) and (b), while the red region and the green region cover triangles that are seen by them exclusively (best view in color)}
    \label{fig:data_gen}
\end{figure}

\noindent\textbf{Triplet sampling using SfM.} As we have shown in the network architecture, the training data is composed of image triplets $(\mathcal{I}_a, \mathcal{I}_p, \mathcal{I}_n)$. Manual annotation for such a large quantity of training triplets is unrealistic. As is observed in~\cite{radenovic2016cnn}, these triplets can be generated from SfM by computing the ratio of shared 3D tracks (which we refer to as \emph{common track ratio}) between views in a fully automatic manner. 
Specifically, suppose $\mathcal{P}(i)$ is the set that contains all the 3D tracks that are observed by image $i$, then the \emph{common track ratio} between the image pair $(\mathcal{I}_i, \mathcal{I}_j)$ is defined as the average of two ratio numbers:

\begin{equation}\label{eq:sfm_overlap}
\mathcal{CT}_{ij} = \mathcal{CT}_{ji} = \mathbf{Ave} (\frac{|\mathcal{P}(i) \cap \mathcal{P}(j) |}{|\mathcal{P}(i)|}, \frac{|\mathcal{P}(i) \cap \mathcal{P}(j) |}{| \mathcal{P}(j)|})
\end{equation}
where the average function $\mathbf{Ave}(a,b) = \sqrt{a \cdot b}$ is the geometric mean. Though other mean functions can be used, we did not observe substantial difference.

\noindent\textbf{Triplet sampling using surface reconstruction.} However, the above sampling method has several drawbacks. First, the generalization power would be limited by the ability of local feature matching. As Fig.~\ref{fig:net_arch}(b) shows, if a pair of matched images possess a large view angle change that exceeds the matching ability of SIFT ($>30^{\circ}$), this pair of images would be regarded as unmatched since few common tracks would exist. Ideally, a good retrieval algorithm should consider all geometrically overlapping pairs and get rid of this limitation.
Second, hard samples as shown in Fig.~\ref{fig:tri_batch} are helpful in matchable image retrieval, in which the triplet images are from the same scene with similar context information. But hard samples cannot be obtained from sampling using SfM, in which negative samples are constrained to be selected from two non-overlapping scenes~\cite{radenovic2016cnn}, since a small ratio of shared tracks does not represent a small overlapping area.

Thus, we combine mesh model re-projection with SfM track overlaps to obtain training triplets.
As shown in Fig.~\ref{fig:data_gen}, we use triangulated mesh models to pinpoint accurate overlap regions between image pairs, which is similar to~\cite{shen2016color}. 
The essence is to project triangular meshes with high level-of-details (LoD) through camera projection matrices registered in SfM.
Similar to \emph{common track ratio}, we define \emph{mesh overlap ratio} between the $(\mathcal{I}_i, \mathcal{I}_j)$ image pair as

\begin{equation}\label{eq:mesh_overlap}
\mathcal{MO}_{ij} = \mathcal{MO}_{ji} = \mathbf{Ave} (\frac{|\mathcal{T}(i) \cap \mathcal{T}(j) |}{|\mathcal{T}(i)|}, \frac{|\mathcal{T}(i) \cap \mathcal{T}(j) |}{| \mathcal{T}(j)|})
\end{equation}
where $\mathcal{T}(i)$ is the set containing all the triangles that are seen by the corresponding camera of image $i$, and $\mathbf{Ave}$ is the same as in Equation~\ref{eq:sfm_overlap} which considers relative scale of image pairs. 
$\mathcal{CT}_{ij}$ and $\mathcal{MO}_{ij}$ are both in the range of $\left[0,1\right]$.

To get a consistent overlap measurement, $\mathcal{CT}_{ij}$ and $\mathcal{MO}_{ij}$ should be carefully merged. The magnitude of $\mathcal{MO}_{ij}$ is usually larger than that of $\mathcal{CT}_{ij}$ in practice. 
We take a SfM-overlap-first scheme to ensure the completeness of positive samples. Namely, the \emph{combining overlap ratio} $\mathcal{CO}$ is defined as

\begin{equation}
\mathcal{CO}_{ij} =
\left\{
\begin{array}{ll}
1  & \mbox{if } \mathcal{CT}_{ij} \geq t_{sfm} \\
\mathcal{MO}_{ij} & \mbox{otherwise }
\end{array}
\right.
\end{equation}

In this work, we fix $t_{sfm}$ to be 0.2 as is used in~\cite{radenovic2016cnn}. An image $\mathcal{I}_j$ is a \textbf{\emph{strong positive}} to the anchor image $\mathcal{I}_i$ if $ \mathcal{CO}_{ij} \in [t_{s1}, t_{s2}] (=[0.5, 1.0])$, and a \textbf{\emph{weak positive}} if $ \mathcal{CO}_{ij} \in [t_{w1}, t_{w2}] (=[0.05, 0.2])$, leaving a safe margin between strong and weak positives. Moreover, the corresponding masks generated by mesh re-projection enable a more accurate computation of the loss term, which will be detailed in the next section.

\subsection{Learning With Batched Hard Mining}\label{sec:triplet_loss}
\noindent\textbf{Triplet Loss.}
The original idea of triplet loss~\cite{wang2014learning,schroff2015facenet} is to push the positive distance $D_+ = D(f(\mathcal{I}_a), f(\mathcal{I}_p))$ far apart from the negative distance $D_- = D(f(\mathcal{I}_a), f(\mathcal{I}_n))$ to a certain margin $\alpha$, formally known as (where $[x]_+ = \max(x, 0)$)

\begin{equation}
\mathcal{L}_{tl}(\mathcal{I}_a,\mathcal{I}_p,\mathcal{I}_n) = [D_+ + \alpha - D_-]_+ 
\end{equation}

\noindent\textbf{Anchor swap.} For symmetric distance measurements like the one in matchable retrieval, the sample space can be halved by introducing in-triplet hard negative mining~\cite{balntaslearning}, which also considers the distance between the positive and the negative $D'_- = D(f(\mathcal{I}_p), f(\mathcal{I}_n))$

\begin{equation}
\mathcal{L}_{as}(\mathcal{I}_a,\mathcal{I}_p,\mathcal{I}_n) = [D_+ + \alpha - min(D_-, D'_-)]_+ 
\end{equation}

\noindent\textbf{Mask triplet loss.} Beyond the similar/dissimilar relations in particular object retrieval, more accurate overlap correspondences can be pin-pointed from the training data generation pipeline described in Section~\ref{sec:data_generation}. Using the groundtruth masks associated with matched image pairs, we propose a new loss termed as \textit{Mask Triplet Loss}

\begin{equation}\label{eq:tle}
\mathcal{L}_{mtl}(\mathcal{I}_a,\mathcal{I}_p,\mathcal{I}_n) = [D_{+}^{\star} - \beta]_+ + \lambda[D_+ + \alpha - min(D_-, D'_-)]_+
\end{equation}
where $D_{+}^{\star} = D(f(\mathcal{I}_a) \odot M(\mathcal{I}_a,\mathcal{I}_p), f(\mathcal{I}_p) \odot M(\mathcal{I}_p,\mathcal{I}_a))$. $\{M(\mathcal{I}_a,\mathcal{I}_p),  M(\mathcal{I}_p,\mathcal{I}_a)\}$ represents a pair of corresponding masks generated by mesh re-projection, and $\odot$ is the masking operation applied on feature maps from CNNs. 
In practice, we use the down-sampled corresponding region maps between the positive image pair $(\mathcal{I}_a,\mathcal{I}_p)$ as a binary filter for pooling operation. The first term in Equation~\ref{eq:tle} penalizes the difference between the masked regions of positive pairs, with a soft margin $\beta$ to prevent overfitting~\cite{lin2017deephash}, while the second term is the triplet loss with anchor swap.
$\beta$ and $\lambda$ are set to 0.1 and 0.5 respectively. 
We have found that the proposed mask triplet loss greatly accelerates the training process since it finds the accurate regions for loss computation.

\noindent\textbf{Batched hard mining.} Since the sample complexity is cubic in the number of images, which is infeasible to iterate over, triplet sampling is vital to ensure the fast convergence of the model. Therefore, a mining strategy~\cite{schroff2015facenet} should be carefully designed to select the proper triplets. Too hard triplets would result in the collapse of the model and too easy triplets would produce no loss and slow down the training process.
Inspired by the previous works used in local descriptor learning, such as structured loss~\cite{song2016deep,luo2018geodesc}, we propose a batched triplet mining strategy suitable for this task which utilizes the fine-grained overlap measurement as defined in Section~\ref{sec:data_generation}.

\begin{figure}[t]
    \centering 
    \includegraphics[width=0.8  \textwidth]{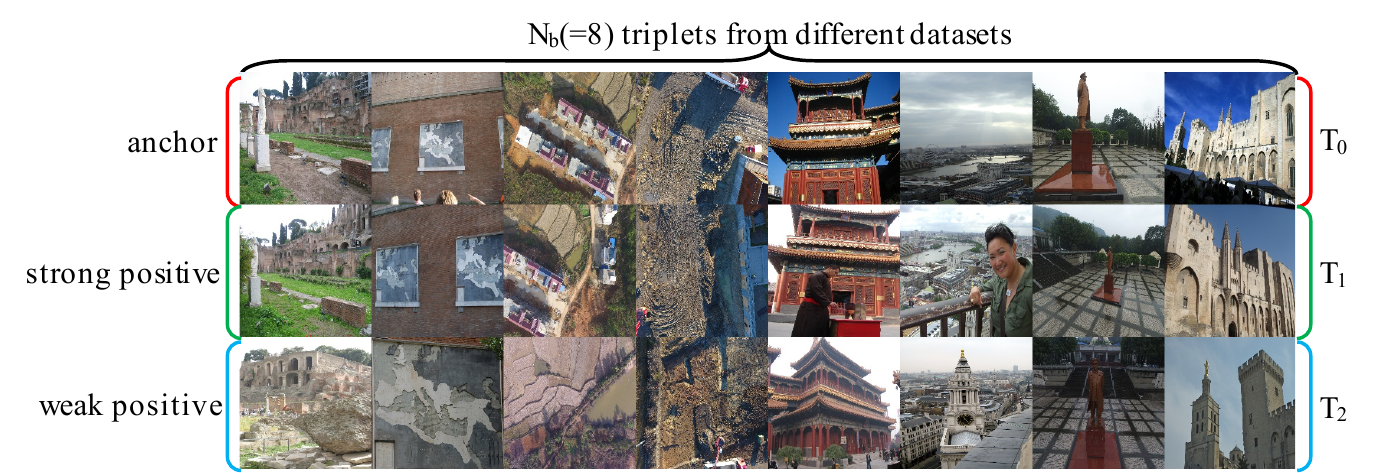}
    \caption{The batched triplets loss formulation}
    \label{fig:tri_batch}
\end{figure}

As shown in Fig.~\ref{fig:tri_batch}, each batch forms a matrix $T$ of size $(3, N_b)$ where $N_b$ is the batch size. Each triplet in the batch comes from a different dataset thus row-wise every pair of images is a \textbf{\emph{negative}} pair. Each column itself forms a \textbf{\emph{hard triplet sample}} meaning that the second row is more similar to anchors (the first row) than the third row, measured by the overlap ratio defined in Section~\ref{sec:data_generation}. We call the second row \textbf{\emph{strong positive}} and the third row \textbf{\emph{weak positive}}. 
The total loss is of three parts: 1) easy loss composed by (anchor, strong positive, negative), 2) weak loss composed by (anchor, weak positive, negative), 3) hard loss composed by (anchor, strong positive, weak positive),  written as follows:

\begin{small}
\begin{equation}\label{eq:batch_tle}
\begin{split}
\mathcal{L}_{T} & = \mathcal{L}_{easy} + \lambda_1 \mathcal{L}_{weak} + \lambda_2 \mathcal{L}_{hard} \\
& = \frac{1}{3N_b (N_{b}-1)}\sum_{i=0}^{N_b-1} \sum_{j=0, j\neq i}^{N_b-1} \sum_{k=0}^{2} [\mathcal{L}_{mtl}(T_{0i},T_{1i},T_{kj}) + \lambda_1 \mathcal{L}_{mtl}(T_{0i},T_{2i},T_{kj})]\\
&  +  \frac{\lambda_2}{N_b}\sum_{i=0}^{N_b-1} \mathcal{L}_{mtl}(T_{0i},T_{1i},T_{2i})
\end{split}
\end{equation}
\end{small}
With this batched loss formulation, the equivalent batch size can be enlarged by an order of magnitude from $O(N_b)$ to $O(N_b^2)$, which makes the training process much more effective. In practice, we set the loss weights $\lambda_1, \lambda_2$ to 1.

\noindent\textbf{Offline mining with adaptive margins.}
Hard negatives are generated offline by mesh re-projection (discussed in Section~\ref{sec:data_generation}). As mentioned in \cite{schroff2015facenet}, we also observe that using hard negatives in the early training process can harm the performance and collapse the model. Therefore, we use adaptive margins, where we set a smaller margin for hard samples to stabilize the training process. We set $\alpha = 1$ for easy triplets, $\alpha = 0.5$ for weak triplets, and $\alpha = 0.5$ for hard triplets.

\subsection{Pre-Matching Regional Code (PRC)}\label{sec:PRC}
Since matchable image retrieval needs fine-grained discrimination of overlap, it is crucial to exploit regional information. R-MAC~\cite{tolias2015particular} provides good insights to tackle this issue. R-MAC samples square regions on activations at different scales, then applies MAC~\cite{azizpour2015generic} on those square regions to get regional vectors which are then combined into a single image vector by summing and L2-normalization.
However, the mixed regional information may weaken its expressive power. 
In this work, we propose \emph{pre-matching regional code (PRC)}, an feature aggregation method towards regional information coding based on~\cite{tolias2015particular,razavian2016visual}.

Generally, PRC can be combined with any pooling operations, such as L2 pooling, average pooling or max pooling. 
We use PRC with max pooling due to its translation invariance~\cite{tolias2015particular,gordo2017end}, which is termed PR-MAC.
We first sample square regions and generate regional vectors as in R-MAC. Instead of simply summing up all the regional vectors, PR-MAC does pre-matching on regional vectors and aggregates the sub-matching result. 
Formally, for an image pair $(\mathcal{I}_Q, \mathcal{I}_T)$ associated with regional vector sets $\{\mathcal{R}_{Q}\}$ and $\{\mathcal{R}_{T}\}$, we first obtain

\begin{equation}\label{eq:prmac}
D_T(\mathcal{R}_{Q, i}) = \min_j\{\norm{\mathcal{R}_{Q, i}, \mathcal{R}_{T, j}}_2\}
\end{equation}
as the minimum distance between a regional vector $\mathcal{R}_{Q, i}$ for the query image $\mathcal{I}_Q$, and the regional vector set $\{\mathcal{R}_T\}$ for the target image $\mathcal{I}_T$. Then we calculate
\begin{equation}\label{eq:10}
D(\mathcal{I}_Q, \mathcal{I}_T) = ||\sum_i(D_T(\mathcal{R}_{Q, i}))||_2
\end{equation}
to represent the final distance between a pair of images. As an interpretation, PRC conducts pre-matching to find the best match for each region of the query, and computes the similarity considering the matchability of each region. As is demonstrated in extensive experiments, PRC outperforms R-MAC in both object image retrieval and matchable image retrieval.

\noindent\textbf{Discussions on efficiency and comparison with R-MAC.} PRC has the computational complexity of $O(k^2)$ where $k$ is the number of regional vectors, which is higher than that of R-MAC. We have improved the efficiency in two ways. First, the PRC is applied on the feature map level as in R-MAC instead of on the costly image patch level~\cite{razavian2016visual}. Second, PRC can be applied on a shortlist (Top-200) as a re-ranking method~\cite{iscen2017efficient}. 
We also compare PRC with approximate max-pooling localization (AML)~\cite{tolias2015particular}, which replaces the sum operation in Equation~\ref{eq:10} with $\arg\max$.

\section{Experiments}\label{sec:exp}

\noindent\textbf{Implementation details.}
We use TensorFlow~\cite{abadi2016tensorflow} to train CNNs on resized $224 \times 224$ images with random contrast and color perturbation.
Various methods of vocabulary tree with advanced techniques~\cite{nister2006scalable,jegou2008hamming,li2015pairwise} are implemented in C++ with multi-threading and SIFT features from VLFeat~\cite{vedaldi2010vlfeat}. Each image has 10k SIFT features on average. We use stochastic gradient descent (SGD) solver with a momentum of 0.9 and a weight decay of 0.0001. The base learning rate is 0.002 and exponentially decayed to 0.9 of the previous one for every 10k steps.
All benchmarks are conducted on single Nvidia GeForce GTX 1080.

\noindent\textbf{Evaluation protocol.} 
We use mean Average Precision (mAP) to measure the performance. We only keep a smaller rank list of size $k$ for each query and measure mAP@k, as only the first fewer candidate matches matter in SfM. Instead of searching the same scene dataset, each image is queried against all 9,368 test images to increase the retrieval difficulty. The ground truth overlap rank list is generated as in Section~\ref{sec:data_generation}. We evaluate the case when $t_{pos} = 0.5$, which results in 317,090 ground-truth match pairs. It provides a challenging benchmark whose images have large scale and perspective changes unlimited by SfM results.

\subsection{Distinctiveness of Matchable Image Retrieval}

\begin{table}[t]
\centering
\caption{We use 896$\times$896 images for CNN-based methods. Dimensionality are reduced to 512 using PCA (or learned whitening $L_W$ computed on the dataset in [11]), computed with an independent dataset with 50k images. QE means weighted query expansion (with top-10 results weighted by the distance), Holiday is not applicable for QE because queries have less than 10 ground-truth images. R-MAC and PR-MAC are used with two scales of 5 (=1+4) regional vectors.}
\label{tab:object_retrieval}
\tiny
    \resizebox{\textwidth}{!}{  
\begin{tabular}{|l|c|c|c|c|c|c|}
\hline
  \textbf{(Dim=512 for all CNN methods)      }                            & \multicolumn{1}{l|}{\textbf{GL3D (mAP@100)}} & \multicolumn{1}{l|}{\textbf{Oxford5k}} & \multicolumn{1}{l|}{\textbf{Paris6k}} & \textbf{Holiday (top-10)} & \textbf{INSTRE} \\ \hline \hline
VocabTree~\cite{nister2006scalable} (depth = 6, branch = 8)                             & 0.599                     & 0.448                         & 0.531                        &    0.549      & -      \\ 
VocabTree + HE + WGC~\cite{jegou2008hamming}                    & 0.689                     & 0.547                         & -                            &    0.746        & -     \\ \hline
siaMAC (VGG) + MAC~\cite{radenovic2016cnn}              &    0.518                &  0.731                          & 0.785                        &    0.723    & 0.296         \\ 
siaMAC (VGG) + R-MAC                         &     0.542               &  0.770                          & 0.821                       &   0.762          & 0.313    \\
siaMAC (VGG) + R-MAC ($L_W$)                         &   0.553                 &   0.779                         & 0.810                       &  0.767   & -           \\  
siaMAC (VGG) + PR-MAC                        & 0.617                     & 0.786                         &  0.832                     &  0.782          & 0.389    \\ 
siaMAC (VGG) + PR-MAC  + QE                      & 0.654                    &  0.830                       & 0.874                        &     -       &0.588    \\ \hline
GoogLeNet + R-MAC + TL &    0.636                    &   0.711                       &       0.794                 &    0.821      & 0.243      \\ 
GoogLeNet + PR-MAC + TL &  0.708                     & 0.737                         &0.813                        &     0.825    &0.306     \\ 
GoogLeNet + PR-MAC + TL + QE &  0.721                     & 0.781                        & 0.855                       &   -     & 0.504       \\
GoogLeNet + R-MAC + MTL                         & 0.638                     & 0.721                        & 0.799                       &    0.824       &-  \\  
GoogLeNet + PR-MAC + MTL                         & 0.711                     & 0.740                       & 0.816                       &   0.841          &-  \\
GoogLeNet + PR-MAC + MTL + QE                         & 0.722                     & 0.789                        & 0.862                       &    -    &-       \\ \hline
\end{tabular}
}
\end{table}

We first demonstrate the intrinsic difference of object retrieval and geometric overlap retrieval, by comparing vocabulary tree, which is extensively used in practical SfM systems~\cite{moulon2013global,sweeney2015optimizing,schonberger2016structure}, and various deep models on GL3D, Oxford5k, Paris6k and INSTRE~\cite{wang2015instre}.
Table~\ref{tab:object_retrieval} shows that siaMAC~\cite{radenovic2016cnn} achieves superior performance on object retrieval tasks but fails to beat even the naive vocabulary tree and our method on the GL3D dataset.
This partially explains the prevalence of vocabulary tree in SfM, and shows that without proper care CNNs do not generalize well on the fine-grained matchable image retrieval problem.

\subsection{Experiments for Matchable Image Retrieval}

Below we give thorough evaluations on GL3D in the context of matchable image retrieval. If not explicitly specified, the CNN methods are tested on $896\times896$ images with PCA whitening and reduced feature dimensionality of 256. Different from Table~\ref{tab:object_retrieval}, we use three scales of 35 (=1+9+25)  region vectors for R-MAC and PR-MAC to demonstrate the best performance.  As Table~\ref{tab:net_perf} shows, the proposed method outperforms all the others.

\begin{table}[t]
\renewcommand\arraystretch{1.2}
    \centering
    \caption{Comparison of different approaches on GL3D. For deep methods, the images are down-sampled to $896\times896$. For vocabulary tree, local descriptors are extracted from full-size $4000\times3000$ images. The time measurement does not count index building for BoW models. The approaches marked by $\diamond$ are baseline models without being fine-tuned on retrieval data. The running time marked with $*$ is evaluated on authors' public code with Matlab or Caffe, and thus may not be comparable.}
    \resizebox{\textwidth}{!}{  
        \tiny
        \begin{tabular}{|l|l|c|c|c|c|c|}
            \hline
            \multicolumn{2}{|c|}{\multirow{2}{*} {\textbf{Approach}}} & \multicolumn{2}{c|}{\textbf{GL3D}} & \textbf{Time} & \multirow{2}{*}{\textbf{Net. type}} & \multirow{2}{*}{\textbf{Dimension}}  \\ \cline{3-5} 
            \multicolumn{2}{|c|}{}                                                 & \textbf{mAP@100}     & \textbf{mAP@200} & \textbf{(min)} &                                                       &                                                                                \\ \hline \hline
            \multirow{22}{*}{\rotatebox{90}{\emph{CNN-based methods}}}   
            & Raw + MAC $\diamond$ & 0.478 & 0.487 & 11.5 & \multirow{9}{*}{VGG-16} & \multirow{5}{*}{512}  \\ \cline{2-5} 
            & SiaMAC~\cite{radenovic2016cnn} + MAC & 0.519 & 0.527 & \multirow{2}{*}{22.6*} & &                                                                                                                                   \\ \cline{2-4}
            & SiaMAC~\cite{radenovic2016cnn} + R-MAC ~\cite{tolias2015particular} &  0.629          &   0.654 &  &                       &                                                                                                                                        \\ \cline{2-5}
            
            & SiaMAC~\cite{radenovic2016cnn} + R-MAC + diffusion~\cite{iscen2017efficient} &  0.569          &   0.598 & 60.5*                        &                                                  &                                                                                      \\ \cline{2-5}
                        
            & SiaMAC~\cite{radenovic2016cnn} + PR-MAC &  0.662          &   0.686 &  60.9*                        &                                                  &                                                                                      \\ \cline{2-5} \cline{7-7}
            & NetVLAD~\cite{arandjelovic2016netvlad} & 0.641 & 0.649 & 28.0* & & \multirow{4}{*}{256}  \\ \cline{2-5}  
                        & Ours + TL + MAC & 0.627 & 0.631 & 9.5 &  &     \\ \cline{2-5}

            & Ours + TL + R-MAC & 0.681 & 0.698 & \multirow{2}{*}{11.5} &  &     \\ \cline{2-4}
            & Ours + MTL + R-MAC & 0.691 & 0.707 &  & &  \\ \cline{2-5}
            & Ours + MTL + PR-MAC & 0.724 & 0.731 & 12.3 &  &   \\ \cline{2-7}
            & Fine-tuned + ROI + R-MAC~\cite{gordo2017end} &  0.616          &   0.629 & 12.6* & ResNet101 & 2048             \\ \cline{2-2}\cline{2-7}         
            & Raw + MAC $\diamond$ &  0.598 & 0.603 & \multirow{5}{*}{3.2} & \multirow{9}{*}{GoogLeNet}                      & \multirow{9}{*}{256}                                                                                                           \\ \cline{2-4} 
            & Ours + TL + MAC & 0.625       & 0.638 &                  &                         &                                                                            \\ \cline{2-4} 
            & Ours + MTL + MAC & 0.652       & 0.663 &             &                         &                                                                            \\ \cline{2-4} 
            & Ours + MTL + SPoC~\cite{babenko2015aggregating} & 0.689       & 0.705    &  &   &                                                                             \\ \cline{2-4} 
            & Ours + MTL + CRoW~\cite{kalantidis2015cross} &   0.673          &   0.698                 &  &                          &                                                                                    \\ \cline{2-5} 
            & Ours + MTL + R-MAC~\cite{tolias2015particular} & 0.702       & 0.715 & 5.4 &         &                                                                                                      \\ \cline{2-5} 
            & Ours + MTL + AML~\cite{tolias2015particular} & 0.630       & 0.637 & 7.2 &         &                                                                                                      \\  \cline{2-5}
            & Ours + MTL + PR-MAC (Top-200)  & 0.722       & 0.743 & 5.5 &  &                                                                                                      \\  \cline{2-5}
            & Ours + MTL + PR-MAC  & \textbf{0.734}       & \textbf{0.758} & 8.5 &  &                                                                                                      \\  \cline{2-4}\cline{5-7} 

            \hline
            
            \multirow{5}{*}{\rotatebox{90}{\emph{VocTree}}} & VocabTree~\cite{nister2006scalable} & 0.599       & 0.614 & 44 & \multirow{5}{*}{-}                       & \multirow{5}{*}{2395371}                                                                        \\ \cline{2-5} 
            & VocabTree + HE~\cite{jegou2008hamming} & 0.601       & 0.615 & 726 &                   &                                                                              \\ \cline{2-5} 
            & VocabTree + WGC~\cite{jegou2008hamming} & 0.676       & 0.688 & 144 &                    &                                                                              \\ \cline{2-5} 
            & VocabTree + PGM~\cite{li2015pairwise} & 0.641       & 0.643 & 173 & &                                                                             \\ \cline{2-5} 
            & VocabTree + HE + WGC~\cite{jegou2008hamming} & 0.689           & 0.703 & 820 & &                                                                                               \\ \hline
        \end{tabular}
    }
    \label{tab:net_perf}
\end{table}

\noindent\textbf{Effect of using hard samples.} Using hard samples is the main benefit brought by our ground-truth generation method (mesh re-projection). Without hard samples, the mAP@200 of our best model drops from 0.758 to 0.717.

\noindent\textbf{Effect of  triplet loss.} We compare the performance training with triplet loss (+TL) and the proposed mask triplet loss (+MTL). MTL and TL deliver similar performance after convergence, as shown in Table~\ref{tab:net_perf}, yet it is observed that MTL converges much faster than TL.

\begin{figure}[t]
    \centering 
    \includegraphics[width=\textwidth]{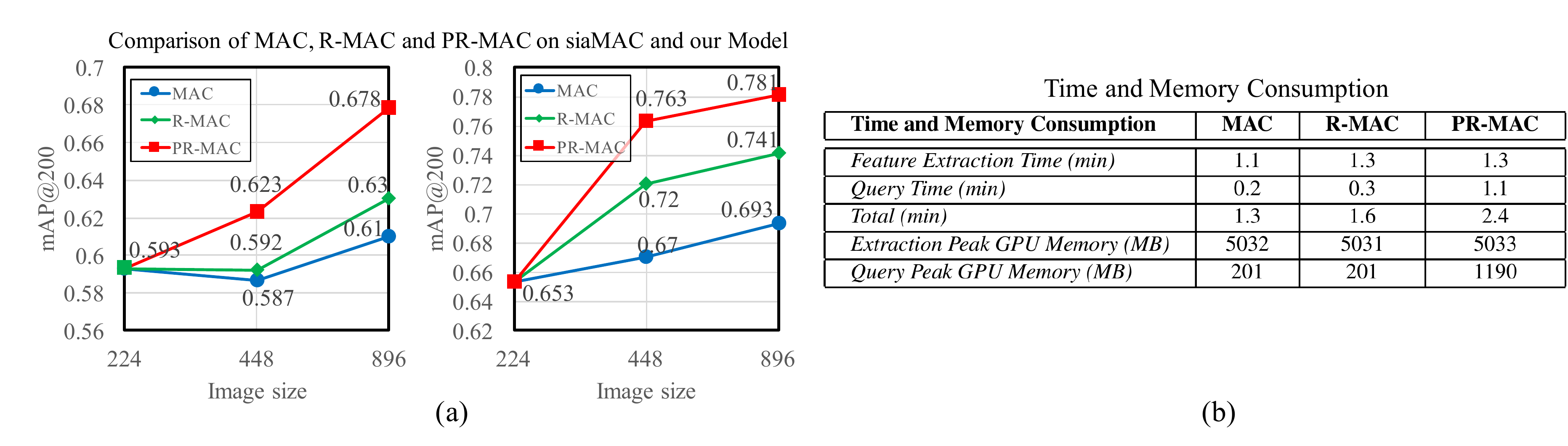} 
    \caption{(a) Comparisons of different aggregation methods. Left: siaMAC model~\cite{radenovic2016cnn}; Right: our fine-tuned model. (b) The time and GPU peak memory consumptions for MAC, R-MAC and PR-MAC during feature extraction and query, carried out on a smaller test dataset from GL3D.}
    \label{fig:rmac_compare}
\end{figure}

\noindent\textbf{Effect of PRC feature aggregation.}
Naturally, images of higher resolution provide richer information and are more likely to deliver better performance. 
To demonstrate that PRC can exploit information not merely from higher resolutions, we compare PR-MAC with MAC and R-MAC for different image sizes. 
As image size increases, PR-MAC consistently outperforms MAC and R-MAC with both siaMAC model~\cite{radenovic2016cnn} (Fig.~\ref{fig:rmac_compare}(a), left) and our fine-tuned model (Fig.~\ref{fig:rmac_compare}(a), right), indicating the versatility of PRC. Moreover, unlike the results in~\cite{radenovic2016cnn} where MAC and R-MAC deliver comparable improvements on object image retrieval, it shows that R-MAC is notably better than MAC in matchable image retrieval, which again demonstrates the difference between two tasks and the necessity of exploiting regional information. We have also found that manifold diffusion method~\cite{iscen2017efficient} and approximate max-pooling localization (AML) in R-MAC~\cite{tolias2015particular} do not work very well on matchable image retrieval, as shown in Table~\ref{tab:net_perf}.

\noindent\textbf{Efficiency.} As shown in Table~\ref{tab:net_perf}, our best model is able to surpass above BoW models regarding both accuracy and efficiency.
Furthermore, Fig.~\ref{fig:rmac_compare}(b) compares the computation time and peak memory for MAC, R-MAC and PR-MAC. 
The higher complexity of PRC can be alleviated to some extent by using PRC as a re-ranking method. 
For example, by applying R-MAC on our best model and then re-ranking the Top-200 candidates with PR-MAC, the mAP@200 score on GL3D increases from 0.715 to 0.743.
Generally, the increase for PR-MAC is due to more I/O operations and the fine-grained matching. However, it still achieves good trade-off to apply PR-MAC for SfM where accuracy is more concerned. 

\subsection{Integration of Matchable Image Retrieval with SfM}

\begin{figure}[t]
    \includegraphics[width=\textwidth]{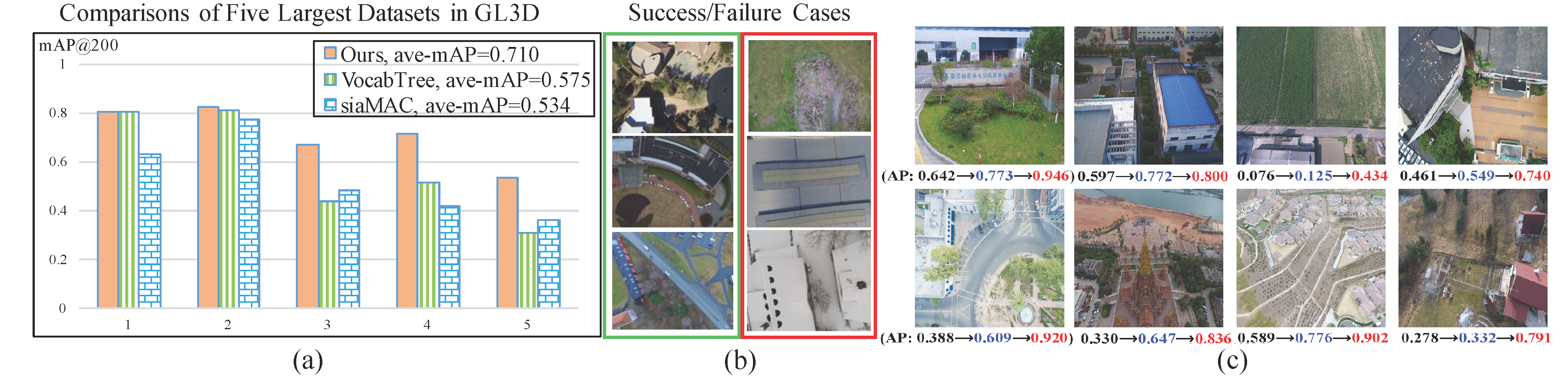}
    \caption{(a) Comparisons of different methods on the five largest scenes in the GL3D dataset. (b) Our method works better on the datasets in the green frame while vocabulary tree does better on the datasets in the red frame. (c) Average precision (AP) for example queries, AP from left to right: \textcolor{black}{siaMAC}$\rightarrow$\textcolor{blue}{VocabTree}$\rightarrow$\textcolor{red}{Ours}.}
    \label{fig:exps}
\end{figure}
\noindent\textbf{Retrieval performance per scene.} Since SfM relies on retrieving matchable images on each independent scene, we extensively evaluate our approach on each of the \testSetNum test sets.
Both our method and vocabulary tree outperform siaMAC and again reflects the gap between object and matchable image retrieval. 
Fig.~\ref{fig:exps}(a) shows the comparisons of the five largest scenes (each more than 400 images) in GL3D.
One observation (Fig.~\ref{fig:exps}(b)) is that our CNN-based method is suitable for datasets with rich textures (the green frame), while vocabulary tree does better on texture-less scenes (the red frame). It indicates that vocabulary tree better encodes very local and detailed information. Performance boost for specific query images is shown in Fig.~\ref{fig:exps}(c).

\begin{table*}[t]
        \centering
        \caption{Evaluation results of different retrieval methods for SfM.}
        \label{tab:eth_benchmark_sfm}
        \resizebox{\textwidth}{!}{  
                \begin{tabular}{ccccccccccc}
                        \Xhline{1pt}
                        & & \textbf{\# Images} & \textbf{\# Registered} & \textbf{\#Pairs-to-Match}& \textbf{\# Sparse Points} & \textbf{\# Observations} & \textbf{Track Length} & \textbf{Reproj. Error} \\ \Xhline{0.7pt}
\textbf{Madrid Metropolis} & \emph{BoW} & 1,344 & 506 & 107,320 & 78,189 & 561K & 7.18 & 0.59px \\
& \emph{siaMAC} & & 433 & 103,355 & 69,192 & 510K & 7.38 & 0.59px \\
& \emph{NetVLAD} & & 467 & 100,876 & 73,724 & 528K & 7.17 & 0.58px \\
& \emph{Ours} & & 494 & 93,238 & 75,339 & 544K & 7.22 & 0.58px \\
\Xhline{0.7pt}
\textbf{Gendarmenmarkt} & \emph{BoW} & 1,463 & 1,067 & 110,476 & 222,557 & 1,441K & 6.47 & 0.67px \\
& \emph{siaMAC} & & 977 & 116,379 & 183,475 & 1,189K & 6.48 & 0.68px \\
& \emph{NetVLAD} & & 1,002 & 105,275 & 201,279 & 1,286K & 6.39 & 0.67px \\
& \emph{Ours} & & 1,049 & 103,091 & 212,745 & 1,349K & 6.34 & 0.66px \\
\Xhline{0.7pt}
\textbf{Tower of London} & \emph{BoW} & 1,576 & 780 & 122,534 & 175,452 & 1,441K & 8.28 & 0.60px \\
& \emph{siaMAC} & & 727 & 120,631 & 160,333 & 1,333K & 8.31 & 0.59px \\
& \emph{NetVLAD} & & 730 & 119,719 & 163,301 & 1,334K & 8.35 & 0.59px \\
& \emph{Ours} & & 740 & 107,044 & 167,426 & 1,386K & 8.28 & 0.59px \\
\Xhline{0.7pt}
\textbf{Alamo} & \emph{BoW} & 2,915 & 972 & 233,040 & 172,553 & 2,084K & 12.08 & 0.63px \\
& \emph{siaMAC} & & 904 & 228,021 &153,483 & 1,948K & 12.69 & 0.64px \\
& \emph{NetVLAD} & & 912 & 218,617 & 158,686 & 1,994K & 12.28 & 0.63px \\
& \emph{Ours} & & 930 & 206,266 & 164,227 & 2,003K & 12.20 & 0.63px \\
\Xhline{0.7pt}
\textbf{Roman Forum} & \emph{BoW} & 2,364 & 1,665 & 179,812 & 357,447 & 2,964K & 8.29 & 0.70px \\
& \emph{siaMAC} & & 1,614 & 185,489 & 320,618 & 2,661K & 8.30 & 0.69px \\
& \emph{NetVLAD} & & 1,635 & 172,870 & 327,778 & 2,702K & 8.27 & 0.70px \\
& \emph{Ours} & & 1,653 & 166,474 & 340,396 & 2,796K & 8.21 & 0.69px \\
\Xhline{0.7pt}
\textbf{ArtsQuad} & \emph{BoW} & 6,514 & 6,037 & 505,593 & 1,354,474 & 9,227K & 6.81 & 0.67px \\
& \emph{siaMAC} & & 5,811 & 496,283 & 1,250,394 & 8,478K & 6.78 & 0.65px & \\
& \emph{Ours} & & 5,887 & 448,500 & 1,290,811 & 8,757K & 6.78 & 0.66px & \\
& \emph{Ours{@}top-115} & & 6,030 & 505,190 & 1,348,521 & 9,122K & 6.82 & 0.66px & \\
\Xhline{0.7pt}
\end{tabular}
        }
\end{table*}

\noindent\textbf{SfM Results.} We conduct SfM experiments on 1DSfM~\cite{wilson_eccv2014_1dsfm} datasets to demonstrate the integration of the proposed method with SfM. The datasets are reconstructed using COLMAP~\cite{schonberger2016structure} with different retrieval methods (BoW,  siaMAC, NetVLAD, and ours), as shown in Table~\ref{tab:eth_benchmark_sfm}. We select top-100 candidates for matching, the default parameter in COLMAP. For CNN methods, the long side of image is resized to 896. siaMAC and NetVLAD are tuned to its best performance (learned whitening, query expansion etc.) as described in their papers. As shown, our method is better than siaMAC and NetVLAD, and comparable with COLMAP-BoW. However, our method generates fewer ($\sim$10\%) match pairs than COLMAP-BoW from the top-100 candidates, indicating more symmetric query results. When fixing the number of pairs to match, e.g., in the last row of table where retrieval is performed at top-115, a similar result as COLMAP-BoW can be obtained. 
Those experiments again validate our observation that there does exist a gap between the matchable and object image retrieval. 
\section{Conclusions}
In this paper, we first differentiate particular object retrieval and matchable image retrieval, and present a large-scale dataset GL3D and a CNN-based method with auto-annotated training data. 
Based on the high-quality fine-grained training data, we utilize the overlap masks obtained from surface reconstruction and develop a batched mask triplet loss to effectively train the network.
Combined with a post-processing method that exploits regional information, this method delivers state-of-the-art performance for matchable image retrieval. \\

\noindent\textbf{Acknowledgment.} This work is supported by T22-603/15N, Hong Kong ITC PSKL12EG02 and the Special Project of International Scientific and Technological Cooperation in Guangzhou Development District (No. 2017GH24).

\bibliographystyle{splncs}
\bibliography{egbib}

\begin{thebibliography}{10}

\bibitem{agarwal2011building}
Agarwal, S., Furukawa, Y., Snavely, N., Simon, I., Curless, B., Seitz, S.M.,
  Szeliski, R.:
\newblock Building rome in a day.
\newblock Communications of the ACM (2011)

\bibitem{moulon2013global}
Moulon, P., Monasse, P., Marlet, R.:
\newblock Global fusion of relative motions for robust, accurate and scalable
  structure from motion.
\newblock In: ICCV. (2013)

\bibitem{sweeney2015optimizing}
Sweeney, C., Sattler, T., Hollerer, T., Turk, M., Pollefeys, M.:
\newblock Optimizing the viewing graph for structure-from-motion.
\newblock In: ICCV. (2015)

\bibitem{schonberger2016structure}
Schonberger, J.L., Frahm, J.M.:
\newblock Structure-from-motion revisited.
\newblock In: CVPR. (2016)

\bibitem{mur2015orb}
Mur-Artal, R., Montiel, J.M.M., Tardos, J.D.:
\newblock Orb-slam: a versatile and accurate monocular slam system.
\newblock IEEE Transactions on Robotics (2015)

\bibitem{sivic2003video}
Sivic, J., Zisserman, A.:
\newblock Video google: A text retrieval approach to object matching in videos.
\newblock In: ICCV. (2003)

\bibitem{nister2006scalable}
Nister, D., Stewenius, H.:
\newblock Scalable recognition with a vocabulary tree.
\newblock In: CVPR. (2006)

\bibitem{kalantidis2015cross}
Kalantidis, Y., Mellina, C., Osindero, S.:
\newblock Cross-dimensional weighting for aggregated deep convolutional
  features.
\newblock In: ECCV Workshop. (2016)

\bibitem{tolias2015particular}
Tolias, G., Sicre, R., J{\'e}gou, H.:
\newblock Particular object retrieval with integral max-pooling of cnn
  activations.
\newblock In: ICLR. (2016)

\bibitem{radenovic2016cnn}
Radenovi{\'c}, F., Tolias, G., Chum, O.:
\newblock Cnn image retrieval learns from bow: Unsupervised fine-tuning with
  hard examples.
\newblock In: ECCV. (2016)

\bibitem{iscen2017efficient}
Iscen, A., Tolias, G., Avrithis, Y., Furon, T., Chum, O.:
\newblock Efficient diffusion on region manifolds: Recovering small objects
  with compact cnn representations.
\newblock In: CVPR. (2017)

\bibitem{havlena2014vocmatch}
Havlena, M., Schindler, K.:
\newblock Vocmatch: Efficient multiview correspondence for structure from
  motion.
\newblock In: ECCV. (2014)

\bibitem{philbin2007object}
Philbin, J., Chum, O., Isard, M., Sivic, J., Zisserman, A.:
\newblock Object retrieval with large vocabularies and fast spatial matching.
\newblock In: CVPR. (2007)

\bibitem{philbin2008lost}
Philbin, J., Chum, O., Isard, M., Sivic, J., Zisserman, A.:
\newblock Lost in quantization: Improving particular object retrieval in large
  scale image databases.
\newblock In: CVPR. (2008)

\bibitem{zhou2017progressive}
Zhou, L., Zhu, S., Shen, T., Wang, J., Fang, T., Quan, L.:
\newblock Progressive large scale-invariant image matching in scale space.
\newblock In: ICCV. (2017)

\bibitem{zhou2018learning}
Zhou, L., Zhu, S., Luo, Z., Shen, T., Zhang, R., Zhen, M., Fang, T., Quan, L.:
\newblock Learning and matching multi-view descriptors for registration of
  point clouds.
\newblock In: ECCV. (2018)

\bibitem{shen2016graph}
Shen, T., Zhu, S., Fang, T., Zhang, R., Quan, L.:
\newblock Graph-based consistent matching for structure-from-motion.
\newblock In: ECCV. (2016)

\bibitem{zhu2017parallel}
Zhu, S., Shen, T., Zhou, L., Zhang, R., Wang, J., Fang, T., Quan, L.:
\newblock Parallel structure from motion from local increment to global
  averaging.
\newblock arXiv preprint arXiv:1702.08601 (2017)

\bibitem{zhu2018very}
Zhu, S., Zhang, R., Zhou, L., Shen, T., Fang, T., Tan, P., Quan, L.:
\newblock Very large-scale global sfm by distributed motion averaging.
\newblock In: CVPR. (2018)

\bibitem{jegou2008hamming}
Jegou, H., Douze, M., Schmid, C.:
\newblock Hamming embedding and weak geometric consistency for large scale
  image search.
\newblock In: ECCV. (2008)

\bibitem{li2015pairwise}
Li, X., Larson, M., Hanjalic, A.:
\newblock Pairwise geometric matching for large-scale object retrieval.
\newblock In: CVPR. (2015)

\bibitem{chum2011total}
Chum, O., Mikulik, A., Perdoch, M., Matas, J.:
\newblock Total recall ii: Query expansion revisited.
\newblock In: CVPR. (2011)

\bibitem{jegou2010aggregating}
J{\'e}gou, H., Douze, M., Schmid, C., P{\'e}rez, P.:
\newblock Aggregating local descriptors into a compact image representation.
\newblock In: CVPR. (2010)

\bibitem{babenko2015aggregating}
Babenko, A., Lempitsky, V.:
\newblock Aggregating local deep features for image retrieval.
\newblock In: ICCV. (2015)

\bibitem{gordo2017end}
Gordo, A., Almazan, J., Revaud, J., Larlus, D.:
\newblock End-to-end learning of deep visual representations for image
  retrieval.
\newblock IJCV (2017)

\bibitem{wang2014learning}
Wang, J., Song, Y., Leung, T., Rosenberg, C., Wang, J., Philbin, J., Chen, B.,
  Wu, Y.:
\newblock Learning fine-grained image similarity with deep ranking.
\newblock In: CVPR. (2014)

\bibitem{schroff2015facenet}
Schroff, F., Kalenichenko, D., Philbin, J.:
\newblock Facenet: A unified embedding for face recognition and clustering.
\newblock In: CVPR. (2015)

\bibitem{melekhovsiamese}
Melekhov, I., Kannala, J., Rahtu, E.:
\newblock Siamese network features for image matching.
\newblock In: ICPR. (2016)

\bibitem{chopra2005learning}
Chopra, S., Hadsell, R., LeCun, Y.:
\newblock Learning a similarity metric discriminatively, with application to
  face verification.
\newblock In: CVPR. (2005)

\bibitem{li2016efficient}
Li, S., Siu, S.Y., Fang, T., Quan, L.:
\newblock Efficient multi-view surface refinement with adaptive resolution
  control.
\newblock In: ECCV. (2016)

\bibitem{ustinova2016learning}
Ustinova, E., Lempitsky, V.:
\newblock Learning deep embeddings with histogram loss.
\newblock In: NIPS. (2016)

\bibitem{simonyan2014very}
Simonyan, K., Zisserman, A.:
\newblock Very deep convolutional networks for large-scale image recognition.
\newblock In: ICLR. (2015)

\bibitem{szegedy2015going}
Szegedy, C., Liu, W., Jia, Y., Sermanet, P., Reed, S., Anguelov, D., Erhan, D.,
  Vanhoucke, V., Rabinovich, A.:
\newblock Going deeper with convolutions.
\newblock In: CVPR. (2015)

\bibitem{shen2016color}
Shen, T., Wang, J., Fang, T., Zhu, S., Quan, L.:
\newblock Color correction for image-based modeling in the large.
\newblock In: ACCV. (2016)

\bibitem{balntaslearning}
Balntas, V., Riba, E., Ponsa, D., Mikolajczyk, K.:
\newblock Learning local feature descriptors with triplets and shallow
  convolutional neural networks.
\newblock In: BMVC. (2016)

\bibitem{lin2017deephash}
Lin, J., Mor{\`e}re, O., Veillard, A., Duan, L.Y., Goh, H., Chandrasekhar, V.:
\newblock Deephash for image instance retrieval: Getting regularization, depth
  and fine-tuning right.
\newblock In: ICMR. (2017)

\bibitem{song2016deep}
Song, H.O., Xiang, Y., Jegelka, S., Savarese, S.:
\newblock Deep metric learning via lifted structured feature embedding.
\newblock In: CVPR. (2016)

\bibitem{luo2018geodesc}
Luo, Z., Shen, T., Zhou, L., Zhu, S., Zhang, R., Yao, Y., Fang, T., Quan, L.:
\newblock Geodesc: Learning local descriptors by integrating geometry
  constraints.
\newblock In: ECCV. (2018)

\bibitem{azizpour2015generic}
Azizpour, H., Sharif~Razavian, A., Sullivan, J., Maki, A., Carlsson, S.:
\newblock From generic to specific deep representations for visual recognition.
\newblock In: CVPR Workshops. (2015)

\bibitem{razavian2016visual}
Razavian, A.S., Sullivan, J., Carlsson, S., Maki, A.:
\newblock Visual instance retrieval with deep convolutional networks.
\newblock ITE Transactions on Media Technology and Applications (2016)

\bibitem{abadi2016tensorflow}
Abadi, M., Barham, P., Chen, J., Chen, Z., Davis, A., Dean, J., Devin, M.,
  Ghemawat, S., Irving, G., Isard, M.,  et~al.:
\newblock Tensorflow: A system for large-scale machine learning.
\newblock In: OSDI. (2016)

\bibitem{vedaldi2010vlfeat}
Vedaldi, A., Fulkerson, B.:
\newblock Vlfeat: An open and portable library of computer vision algorithms.
\newblock In: ACM Multimedia. (2010)

\bibitem{wang2015instre}
Wang, S., Jiang, S.:
\newblock Instre: a new benchmark for instance-level object retrieval and
  recognition.
\newblock ACM Transactions on Multimedia Computing, Communications, and
  Applications (TOMM) (2015)

\bibitem{arandjelovic2016netvlad}
Arandjelovic, R., Gronat, P., Torii, A., Pajdla, T., Sivic, J.:
\newblock Netvlad: Cnn architecture for weakly supervised place recognition.
\newblock In: CVPR. (2016)

\bibitem{wilson_eccv2014_1dsfm}
Wilson, K., Snavely, N.:
\newblock Robust global translations with 1dsfm.
\newblock In: ECCV. (2014)

\end{thebibliography}

\end{document}